\documentclass{article}

\usepackage[final]{neurips_2021}

\usepackage[utf8]{inputenc} 
\usepackage[T1]{fontenc}    
\usepackage{hyperref}       
\usepackage{url}            
\usepackage{booktabs}       
\usepackage{amsfonts}       
\usepackage{nicefrac}       
\usepackage{microtype}      
\usepackage{xcolor}         
\usepackage{microtype}
\usepackage{graphicx}
\usepackage{subfigure}
\usepackage{booktabs} 

\usepackage{hyperref}

\definecolor{pastel_blue}{rgb}{0.86, 0.926, 0.984}

\usepackage{adjustbox}
\usepackage[frozencache,cachedir=.]{minted} 
\usepackage{xcolor}
\usepackage{array}
\usepackage{siunitx}
\usepackage{xspace}
\usepackage{ulem}
\usepackage[T1]{fontenc}
\setminted{fontsize=\scriptsize}
\definecolor{bg}{HTML}{282828} 
\usepackage{pmboxdraw}

\newcommand*{\score}[1]{\num[round-mode=places,round-precision=1]{#1}}
\newcommand*{\bleu}[1]{\num[round-mode=places,round-precision=2]{#1}}
\newcommand*{\compacc}[1]{\num[round-mode=places,round-precision=1]{#1}}
\newcommand*{\acc}[1]{\num[round-mode=places,round-precision=3]{#1}}

\def\transcoder{TransCoder\xspace}

\def\deobf{DOBF\xspace}

\def\codebert{CodeBERT\xspace}
\def\graphcodebert{GraphCodeBERT\xspace}
\def\roberta{RoBERTa\xspace}
\newcommand{\resubmition}[1]{\textcolor{black}{#1}}
\newcommand{\newbr}[1]{\textcolor{black}{#1}}

\begin{document}
	
	\title{DOBF: A Deobfuscation Pre-Training Objective for Programming Languages}
	
	
	
	\makeatletter
	\newcommand{\printfnsymbol}[1]{%
		\textsuperscript{\@fnsymbol{#1}}%
	}
	\makeatother
	
	\author{
		Baptiste Roziere\thanks{Equal contribution. The order was determined randomly.}\\
		Facebook AI Research\\ Paris-Dauphine University\\
		broz@fb.com\\
		\And 
		Marie-Anne Lachaux\printfnsymbol{1}\\
		Facebook AI Research\\
		malachaux@fb.com\\
		\And 
		Marc Szafraniec\\
		Facebook AI Research\\
		szafraniec@fb.com\\
		\vspace{-0.5cm}
		\And 
		Guillaume Lample\\
		Facebook AI Research\\
		glample@fb.com\\
		\vspace{-0.5cm}
		%
		%
	}
	
	
	
	
	
	
	
	\maketitle
	\begin{abstract}
		
		Recent advances in self-supervised learning have dramatically improved the state of the art on a wide variety of tasks.  
		However, research in language model pre-training has mostly focused on natural languages, and it is unclear whether models like BERT and its variants provide the best pre-training when applied to other modalities, such as source code.
		In this paper, we introduce a new pre-training objective, DOBF, that leverages the structural aspect of programming languages and pre-trains a model to recover the original version of obfuscated source code.
		We show that models pre-trained with DOBF significantly outperform existing approaches on multiple downstream tasks, providing relative improvements of up to \resubmition{12.2\% in unsupervised code translation, and 5.3\%} in natural language code search.
		Incidentally, we found that our pre-trained model is able to deobfuscate fully obfuscated source files, and to suggest descriptive variable names.
		\vspace{-0.2cm}
	\end{abstract}
	
	\section{Introduction}
	
	Model pre-training with self-supervised methods such as BERT~\cite{devlin2018bert}, \roberta~\cite{liu2019roberta}, XLM~\cite{lample2019cross} or XLNet~\cite{yang2019xlnet}, has become ubiquitous in Natural Language Processing (NLP), and led to significant improvements in many tasks.
	These approaches are based on the Masked Language Modeling (MLM) objective, which consists in randomly masking words from an input text, and training a model to recover the original input.
	In the original approach proposed by \citet{devlin2018bert}, a fraction of selected masked words is replaced by masked tokens, another is replaced by random words, and another remains unchanged.
	Since then, a myriad of studies have proposed to modify the MLM objective, either by masking contiguous spans of text \cite{song2019mass, joshi2020spanbert}, masking named entities and phrases \cite{sun2019ernie}, sampling masked words according to their frequencies \cite{lample2019cross}, replacing words with plausible alternatives \cite{clark2020electra}, etc.
	Overall, most of these pre-training objectives boil down to denoising auto-encoding tasks with different methods to add noise to the input, using arbitrary noise functions.
	In our case, we are interested in pre-training deep learning models for programming languages. As in natural language, pre-training was shown to be effective for source code~\cite{feng2020codebert, roziere2020unsupervised}.
	However, these studies both rely on the original MLM objective proposed by \citet{devlin2018bert}, which was initially designed for natural languages and does not leverage the particular structure of source code.
	We argue that this objective is actually suboptimal in the context of programming languages, and propose a new objective based on deobfuscation of identifier names in source code.
	
	Code obfuscation consists in modifying source code in order to make it harder for humans to understand, or smaller while keeping its behaviour unchanged. In some ancient interpreted languages, name minimization could also reduce the memory usage of the program. Today, it is used to protect intellectual property by preventing people from understanding and modifying the code, to prevent malware detection, and to compress programs (e.g. Javascript code) to reduce network payload sizes.
	Moreover, C compilers discard variable names, and current rule-based and neural-based decompilers generate obfuscated C code with uninformative variable names~\cite{fu2019coda}.
	Obfuscators typically apply several transformations to the code. While some operations can be reversed (e.g. dead code injection), the obfuscation of identifier names\textemdash renaming every variable, method and class with uninformative names\textemdash is irreversible and has a substantial impact on code comprehension~\cite{gellenbeck1991investigation,takang1996effects,lawrie2006s}.
	
\begin{figure*}[t]
		\makebox[\textwidth][c]{\includegraphics[width=1.0\textwidth]{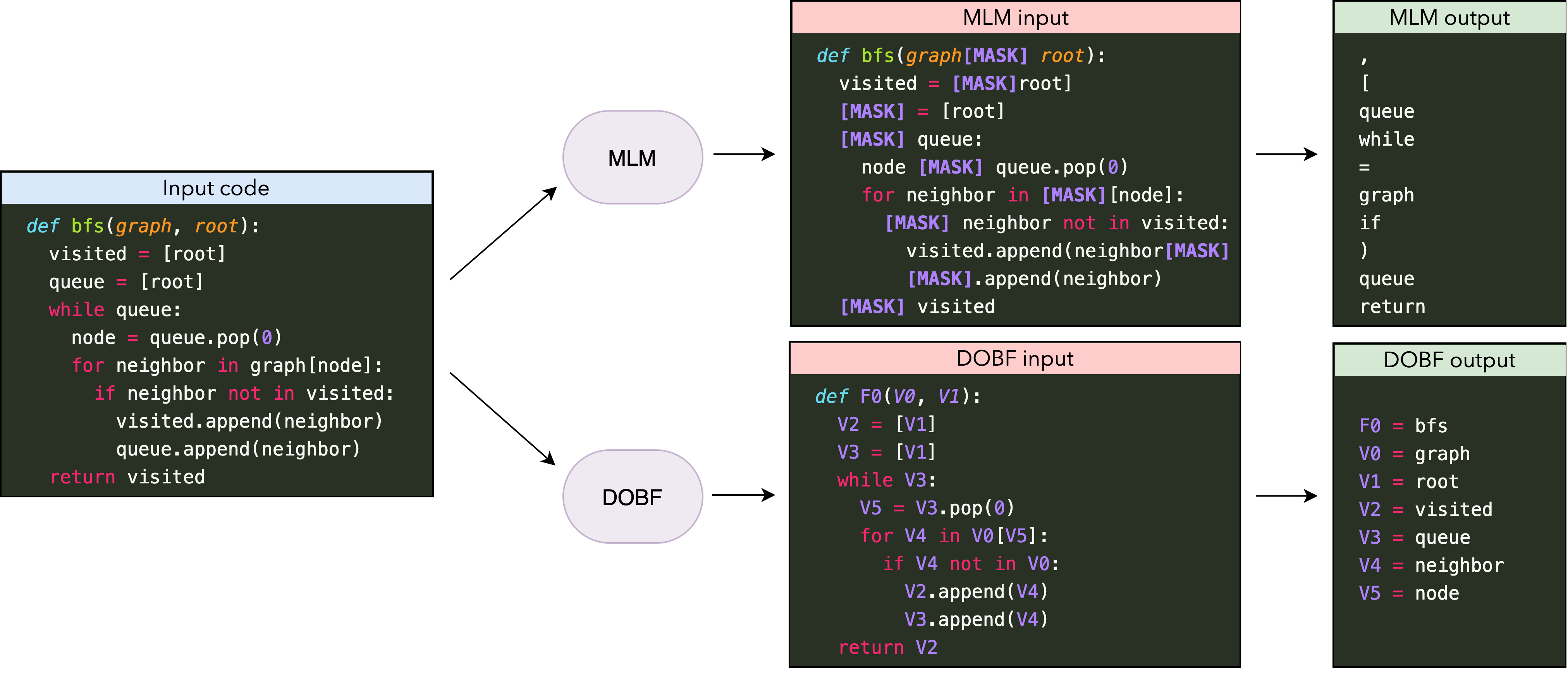}}%
		\caption{\small
			\label{fig:illustration}
			\small
			\textbf{Illustration of the MLM and \deobf objectives. }
			Given an input function, the masked language modeling (MLM) task randomly samples tokens to mask out. With source code, a large fraction of these tokens are related to the language syntax (e.g. commas, parentheses, etc.) that are trivial for the model to predict, and provide a poor training signal.
			Instead, we propose to obfuscate the code by masking the name of functions and variables, and to train the model to recover the original function by deobfuscating the code (\deobf). When a variable is masked out, we mask all occurrences of this variable with the same mask symbol (e.g. all occurrences of ``visited'' are replaced by ``V0'') to prevent the model from copying names. The \deobf objective is more difficult and provides a better learning signal. 
		}
	\end{figure*}

	By analyzing the overall structure of an obfuscated file, an experienced programmer can always, with time, understand the meaning of the obfuscated code. For instance, in the obfuscated example in Figure~\ref{fig:illustration}, one can recognize the function and guess that it implements a breadth-first search algorithm.
	We also expect neural networks, that excel in pattern recognition, to perform well on this task.
	We propose to pre-train a model to revert the obfuscation function, by training a sequence-to-sequence (seq2seq) model to convert obfuscated functions, where names of functions and variables have been replaced by uninformative names, back to their original forms.
	Suggesting proper variable and function names is a difficult task that requires to understand what the program does.
	In the context of source code, it is a more sensible, but also a more difficult task than MLM.
	Indeed, we observe (c.f. Figure~\ref{fig:illustration}) that predicting the content of randomly masked tokens is usually quite simple, as it often boils down to making syntax related predictions (e.g. predicting that was has been masked out is a parenthesis, a semi-column, etc.). These simple predictions actually provide little training signal to the model.
	In practice, MLM also masks out variable names, but if a given variable appears multiple times in a function, it will be easy for the model to simply copy its name from one of the other occurrences.
	Our model does not have this issue, as all occurrences of masked variables are replaced by the same \texttt{VAR\_i} special tokens.

	In this paper, we make the following contributions:
	\begin{itemize}
		\item We present \deobf, a new pre-training objective based on deobfuscation, and show its effectiveness on multiple programming languages.
		\item We show that \deobf significantly outperform MLM (e.g. BERT) on multiple tasks such as code search, code summarization or unsupervised code translation. 
		\resubmition{We show that pre-training methods based on \deobf outperform all existing pre-training methods on all the considered tasks.}
		\item We show that, by design, models pre-trained with \deobf have interesting applications and can be used to understand functions with uninformative identifier names.
		Besides, the model is able to successfully deobfuscate fully obfuscated source files.
	\end{itemize}

	\section{Related work}
	
	
	\paragraph{Masked Language Modeling pre-training.}
	Large pre-trained transformers such as BERT~\cite{devlin2018bert} or \roberta~\cite{liu2019roberta} led to significant improvements in the majority of natural language processing tasks.
	The quality of pre-training mainly comes from the MLM objective (i.e. the cloze task), that allows the model to make predictions by leveraging left and right contexts, unlike causal language modeling (CLM) where the model predictions are only conditioned on previous words.
	In MLM, the model takes as input a sentence and uniformly selects 15\% of its tokens. Of the selected tokens, 80\% are replaced by a special symbol [MASK], 10\% are left unchanged, and the remaining 10\% are replaced by random tokens from the vocabulary. The MLM objective consists in recovering the initial sentence given the corrupted one.
	\citet{lample2019cross} noticed that the masked words are often easy to predict, and proposed to sample the 15\% masked words according to their frequencies instead of uniformly.
	This way, rare words are sampled more often, making the pre-training task more difficult for the model, which results in a better learning signal and faster training.
	\citet{sun2019ernie} also noticed that recovering the tokens masked by MLM is too simple in some contexts (e.g. predicting the two tokens ``Harry Potter'' is much harder than predicting only ``Harry'' if you know the next word is ``Potter''). To address this issue, they proposed to mask phrases and named entities instead of individual tokens. \citet{joshi2020spanbert} and \citet{song2019mass} made a similar observation and proposed to mask random spans of text. They showed that this simple modification improves the performance on many downstream NLP tasks.
	
	\paragraph{Alternative objectives.}
	Other pre-training objectives have been proposed in addition to MLM.
	For instance, \citet{devlin2018bert} also uses the next sentence prediction (NSP) objective, a binary classification task that consists in predicting whether two input sentences follow each other in the original corpus.
	The NSP objective was originally designed to improve the performance on downstream NLP tasks, but recent studies \cite{lample2019cross, liu2019roberta} showed that training MLM on a stream of sentences to leverage longer context, and removing the NSP objective improves the quality of pre-training.
	To improve the sample-efficiency of MLM (where only 15\% of tokens are predicted), Electra~\cite{clark2020electra} proposed to replace (and not mask) some tokens with plausible alternatives, and to train a network to detect the tokens that have been replaced.
	They showed that this new Replaced Token Detection (RTD) objective matches the performance of \roberta while using four times less computational resources.
	\citet{dong2019unified} proposed a model that combines multiple pre-training tasks, including bidirectional, but also left-to-right and right-to-left language modeling objectives.
	\citet{lewis2019bart} also proposed different pre-training objectives, to detect whether input sentences have been permuted, tokens have been deleted or inserted, etc.
	
	
	\paragraph{Code Generation Pre-training.}
	Recent studies showed that pre-training methods developed for natural language processing are also effective for programming languages.
	For instance, \citet{feng2020codebert} proposed \codebert, a \roberta-based model trained on source code using the MLM and RTD objectives. \resubmition{With \graphcodebert~\cite{guo2020graphcodebert}, the MLM objective is complemented by an edge-prediction objective, in which the model predicts edges in the data flow graph to make the model understand the structure of the code.
		In \citet{jain2020contrastive}, a model is trained on javascript code using a contrastive loss ensuring that the representations are robust to some semantic-preserving transformations.} 
	They showed that their model performs well on downstream code generation tasks and outperforms previous pre-training approaches. \citet{kanade2020learning} applied MLM and the next sentence prediction objectives to pre-train models on Python code.
	More recently, \citet{roziere2020unsupervised} applied the unsupervised machine translation principles of \citet{lample2018unsupervised, lample2018phrase} to monolingual source code from GitHub. They showed that the resulting model, \transcoder, was able to translate source code between Python, Java, and C++, in a fully unsupervised way.
	In this paper, we propose to use a code-specific objective to better pre-train models designed to be fine-tuned on code generation tasks: code deobfuscation.
	\resubmition{Machine learning is frequently used on tasks involving programming languages, including code completion~\cite{li2017code,liu2020self,kim2020code,svyatkovskoy2020fast}, bug detection and code repair~\cite{Allamanis2018LearningTR,wang2017dynamic,chen2019sequencer,murali2020industry,tufano2019empirical,tarlow2020learning}, code summarization~\cite{alon2018code2seq,hu2018deep}, clone detection~\cite{wei2017supervised,ain2019systematic,wang2020detecting}, code search~\cite{gu2018deep,cambronero2019deep} and code translation~\cite{chen2018tree,roziere2020unsupervised}. Most of these tasks can benefit from pre-trained models that capture the semantics of the code.}
	
	\paragraph{Code deobfuscation.}
	Empirical studies show that naming conventions and the use of informative identifier names make code more understandable, easier to maintain and lead to fewer bugs~\cite{takang1996effects, liblit2006cognitive, butler2009relating}.
	It motivated other works studying deobfuscation of identifier names and identifier name proposal using n-grams~\cite{allamanis2014learning, allamanis2015suggesting}, probabilistic models~\cite{raychev2015predicting, bichsel2016statistical, vasilescu2017recovering, alon2018general}, and recurrent neural networks~\cite{bavishi2018context2name, lacomis2019dire}. 
	\citet{alon2018general} extract features from Abstract Syntax Tree (AST) paths and train a Conditional Random Field to predict variable and method names, and infer types for several languages.
	DIRE~\cite{lacomis2019dire} uses a commercial decompiler to obtain C code with uninformative identifier names from binaries. They also use AST features, which go through a Graph Neural Network trained jointly with a LSTM model on the sequence of C tokens to retrieve relevant identifier names.
	More recently, \citet{david2020neural} used a transformer together with augmented representations obtained from static analysis to infer procedure names in stripped binary files.
	These models are already used to understand obfuscated and compiled source code.
	However, none of these studies investigated the use of deobfuscation for model pre-training.

	\section{Model}
	
	\subsection{MLM and denoising for Programming Languages}
	
	A countless number of pre-training objectives have been introduced in the literature \cite{devlin2018bert, clark2020electra, lewis2019bart, liu2019roberta, dong2019unified}. Most of them rely on hyper-parameters and seemingly arbitrary decisions (Should we mask individual tokens or spans? Which fraction of them? What do we do with masked out tokens? etc.).
	These choices are typically based on intuition and validated empirically on natural language processing tasks.
	However, source code is much more structured than natural language, which makes predicting masked tokens much easier for programming languages.
	
	The first row in Figure~\ref{fig:illustration} shows an example of input / output for the MLM objective. We can see that the majority of tokens are composed of Python keywords or symbols related to syntax: \texttt{, [ while = if ) return}. These symbols are easy to recover, and a model will quickly learn to predict them with perfect accuracy.
	This effect is accentuated by the verbosity of the language. For instance, we would see significantly more of these tokens in Java.
	Retrieving the obfuscated \texttt{graph} token is also relatively simple: the model only needs to retrieve the most relevant variable in the scope. More generally, retrieving an identifier name is often easy when given its full context, including its definition and usages.
	The denoising-auto-encoding (DAE) objective~\cite{vincent2008extracting}, which trains an encoder-decoder model to retrieve masked token and recover randomly modified input sentences, is quite similar to MLM and the model can also retrieve identifier names easily by finding their definition or usages.
	Overall, we suspect that the MLM objective is too simple in programming languages and we introduce a new objective, \deobf, which encourages the model to learn a deeper understanding of code semantics.
	
	\subsection{Deobfuscation Objective}

	Instead of MLM, we propose a new pre-training objective, \deobf, that leverages the particular structure of programming languages.
	We obfuscate code snippets by replacing class, function and variable names with special tokens, and train a model to recover the original names.
	When an identifier is selected, all of its instances in the code are replaced by the same special token.
	This differs from MLM where the name of a variable can appear multiple times while being masked a single time.
	For instance, in Figure~\ref{fig:illustration}, \deobf will replace the two occurrences of \texttt{node} by the same symbol \texttt{V5}, while MLM will only mask one of these occurrences.
	As a result, the fraction of meaningful tokens masked by the objective is language independent: for more verbose languages (e.g. Java), the less informative syntax-related tokens will not be masked out by the \deobf objective.
	
	Each identifier is replaced with probability $p_{obf} \in [0, 1]$. We ensure that the original input is modified: if no identifier is replaced, we draw a random one to obfuscate.
	When $p_{obf} = 0$, we always obfuscate exactly one random identifier in the input. When $p_{obf} = 1$, we obfuscate all the identifiers defined in the file.
	We ensure that the obfuscated code has the same behavior as the original.
	The second row in Figure~\ref{fig:illustration} shows an example of obfuscated code with $p_{obf} = 1$, where we obfuscate a function \texttt{bfs} which implements a breadth-first search.
	The function \texttt{append} is not obfuscated as it is a standard Python function not defined in the file.
	The model is given the obfuscated code as input and has to restore the original name of each special token \texttt{CLASS\_i}, \texttt{FUNC\_i} and \texttt{VAR\_i}.
	In other words, the model needs to output a dictionary mapping special tokens to their initial values.
	
	Finding informative names for obfuscated identifiers requires the model to learn a deep understanding of code semantics, which is desirable for a pre-training task.
	MLM will mask only some of the occurrences of the identifiers and leave the other ones unchanged so that the model can simply copy identifier names.
	In Figure~\ref{fig:illustration}, with MLM masking, the model can simply notice that a variable named \texttt{queue} is called on the fourth line.
	Since the variable is not defined, the model can easily guess that \texttt{queue} has to be defined on the third line, and infer the value of the corresponding \texttt{[MASK]} token.
	With the deobfuscation objective, the model needs to analyze code patterns and understand the semantics of the variable to infer that, since its elements are popped with \texttt{.pop(0)}, the variable \texttt{V3} implements a queue.
	If its elements were popped with \texttt{.pop()}, our model would name it \texttt{stack} instead of \texttt{queue} (c.f. Figure~\ref{fig:graph_traversal} in the appendix).
	
	
	
	\subsection{Implementation}
	
	Overall, the deobfuscation objective operates like a supervised machine translation objective, where a seq2seq model is trained to map an obfuscated code into a dictionary represented as a sequence of tokens.
	At inference time, the model is able to suggest meaningful class, function and variable names for a piece of code with an arbitrary number of obfuscated identifiers.
	Obfuscated classes, functions, and variables, are replaced with associated special tokens: \texttt{CLASS\_0}~\dots~\texttt{CLASS\_N}, \texttt{FUNC\_0}~\dots~\texttt{FUNC\_N} and \texttt{VAR\_0}~\dots~\texttt{VAR\_N}.
	We serialize the output dictionary as a sequence of tokens where the entries are separated by a delimiter symbol \texttt{|}.
	\footnote{In the obfuscated example given in Figure~\ref{fig:illustration}, the model is trained to generate: \texttt{FUNC\_0 bfs | VAR\_0 graph | VAR\_1 root | VAR\_2 visited | VAR\_3 queue | VAR\_4 neighbor | VAR\_5 node}.}
	
	\section{Experiments}
	
	We train \deobf with the deobfuscation objective. First, we evaluate our model on two straightforward deobfuscation applications. Then, we show its performance on multiple downstream tasks.
	
	\subsection{Deobfuscation}
	We evaluate our model on two applications of the deobfuscation task: when $p_{obf}=0$ (the model has to retrieve a single identifier name), and $p_{obf}=1$ (the model has to retrieve all the identifier names).
	
	\paragraph{Deobfuscating a single identifier}
	When $p_{obf}=0$, only one identifier is obfuscated. In that case, the model has to propose a relevant name for that identifier using the rest of the non-obfuscated file as context.
	It can be used as a tool that suggests relevant variable names.
	Integrated development environments (e.g. PyCharm, VSCode) already perform this task, often using handcrafted rules.
	
	\paragraph{Deobfuscating all identifiers}
	Obfuscators are commonly used to make code smaller and more efficient or to protect it by making it more difficult to understand and reuse.
	They typically apply several transformations, one of them being to replace every identifier name with short and uninformative names (e.g. a, b, c).
	In our work, such a transformation corresponds to obfuscating a file with $p_{obf}=1$.
	To measure our model's ability to revert the obfuscation operation, we evaluate its accuracy when obfuscating all identifier names.
	Another application would be to help understand source code written with uninformative variable names.
	
	\paragraph{Evaluation metric}
	We evaluate the ability of our model to retrieve identifier names from the original non-obfuscated code.
	We report the accuracy, which is the percentage of recovered tokens that exactly match the ground truth.
	Following previous works \cite{allamanis2015suggesting, allamanis2016convolutional, alon2018general, alon2019code2vec}, we also report the \textit{subtoken score}, a more flexible metric which computes the precision, recall, and F1 scores for retrieving the original case-insensitive subtokens.
	Each token is broken into subtokens using uppercase letters for camlCase and underscores for snake\_case.
	For instance, \texttt{decoderAttention} would be considered to be a perfect match for \texttt{decoder\_attention} or \texttt{attentionDecoder}. \texttt{attention} would have a perfect precision but a recall of $0.5$, so a F1 score of 66.7. \texttt{crossAttentionDecoder} would have a perfect recall but a precision of $\frac23$, corresponding to a F1 score of $80.0$. 
	We compute the overall subtoken precision, recall and F1 scores averaged over each file in our validation and test datasets.
	
	
	\subsection{Fine-tuning on downstream tasks}
	\label{sec:finetuning}
	In order to evaluate \deobf as a pre-training model, we fine-tune \deobf on TransCoder and on three tasks from CodeXGLUE~\cite{CodeXGLUE}, a benchmark for programming languages.
	\resubmition{The data, code and models from CodeXGLUE and TransCoder are available respectively under the MIT and the Creative Commons license.}
	We only consider the Java and Python tasks with an encoder in the model architecture for which the training, validation, and test sets are publicly available.
	
	\textbf{CodeXGLUE Clone Detection} This task is a binary classification problem where the model has to predict whether two code snippets are semantically equivalent. It is evaluated using the macro F1 score. The model is composed of a single encoder and a classification layer. An input consists in two snippets of code, which are concatenated before being fed to the model.
	This task is available in Java.
	
	\textbf{CodeXGLUE Code Summarization} Given a code snippet, the model is trained to generate the corresponding documentation in natural language. The architecture is a sequence-to-sequence transformer model evaluated using BLEU score~\cite{bleu}. The dataset includes both Java and Python source code. 
	
	\textbf{CodeXGLUE NL Code Search} Given a code search query in natural language the model has to retrieve the most semantically related code within a collection of code snippets. This is a ranking problem evaluated using the Mean Reciprocal Rank (MRR) metric. The model is composed of two encoders. The natural language query and the code are encoded separately, and we compute the dot product between the first hidden states of the encoders' last layers. This task is available in Python.  
	
	\textbf{TransCoder} TransCoder \cite{roziere2020unsupervised} is an unsupervised machine translation model which translates functions and methods between C++, Java, and Python. A single seq2seq model is trained for all languages. In the original work, TransCoder is pre-trained with MLM, and trained with denoising auto-encoding and back-translation. TransCoder is evaluated using the Computational Accuracy metric, which computes the percentage of correct solutions according to series of unit tests. We only consider a single model output (CA@1), with beam sizes of 1 and 10.
	

	\subsection{Experimental details}
	\label{sec:experimental_details}
	
	\textbf{Model Architecture} We consider a seq2seq model with attention, composed of an encoder and a decoder using a transformer architecture~\cite{vaswani2017attention}. We train models with the same architecture and tokenizer as \codebert{}~\cite{feng2020codebert} and \graphcodebert{}~\cite{guo2020graphcodebert} in order to provide fair comparisons\resubmition{: 12 layers, 12 attention heads and a hidden dimension of 768. } \newbr{We also train a model with the same parameters as \transcoder (see Figure~\ref{tab:codexglue_transcoder} in the Appendix).}
	
	\textbf{Training dataset}
	As in \citet{roziere2020unsupervised}, we use the GitHub public dataset available on Google BigQuery and select all Python and Java files within the projects \resubmition{with licenses authorizing use for research purposes.} 
	Following \citet{lopes2017dejavu} and \citet{allamanis2019adverse}, we remove duplicate files.
	We also ensure that each fork belongs to the same split as its source repository.
	We obfuscate each file and create the corresponding dictionary of masked identifier names, resulting in a parallel (obfuscated file - dictionary) dataset of 19 GB for Python and 26 GB for Java.
	We show some statistics about this dataset in Table~\ref{tab:dataset} \resubmition{in the appendix.}
	For comparison purposes, we apply either the BPE codes used by \citet{roziere2020unsupervised} or by \citet{feng2020codebert}.
	In practice, we train only on files containing less than 2000 tokens, which corresponds to more than 90\% and 80\% of the Java and Python files respectively.
	
	\textbf{Training details}
	We train \deobf to translate obfuscated files into lists of identifier names.
	During \deobf training, we alternate between batches of Java and Python composed of 3000 tokens per GPU. We optimize \deobf with the Adam optimizer \cite{kingma2014adam} and an inverse square-root learning rate scheduler~\cite{vaswani2017attention}. We implement our models in PyTorch \cite{paszke2019pytorch} and train them on 32 V100 GPUs \resubmition{ for eight days}. We use float16 operations to speed up training and to reduce the memory usage of our models. We try different initialization schemes: training from scratch and with a Python-Java MLM model following \citet{roziere2020unsupervised}. We train \deobf with three different obfuscation probability parameters: $p_{obf} \in \{0, 0.5, 1\}$. For each $p_{obf}$ value, we train models with multiple initial learning rates ranging from $10^{-4}$ to $3.10^{-4}$ and select the best one using the average subtoken F1 score computed on the validation dataset.
	
	\textbf{Fine-tuning details}
	Depending on the fine-tuning tasks, we consider different model architectures: seq2seq models with encoder and decoder, architectures with two encoders or a single encoder.
	In all cases, we initialize the encoders of these models with the encoder of \deobf and fine-tune all parameters.
	For fair comparison, we rerun all baselines, and train models with the same architectures, number of GPUs, batch sizes and optimizers.
	For CodeXGLUE, we noticed that the tasks are quite sensitive to the learning rate parameter used during fine-tuning. We perform a grid search on five learning rate parameters ranging from $5.10^{-6}$ to $10^{-4}$ and we select the best parameter on the validation dataset. 
	For TransCoder, we use a learning rate of $10^{-4}$ as in \citet{roziere2020unsupervised} \resubmition{and we train the models for 2 day on 32 Tesla V100 GPUs.}

	\begin{figure}
		\begin{minipage}{0.45\textwidth}
			\begin{minted}{python}
			def FUNC_0(VAR_0, VAR_1):
			VAR_2 = [VAR_1]
			VAR_3 = [VAR_1]
			while VAR_3:
			VAR_4 = VAR_3.pop(0)
			for VAR_5 in VAR_0[VAR_4]:
			if (VAR_5 not in VAR_2):
			VAR_2.add(VAR_5)
			VAR_3.append(VAR_5)
			return VAR_2
			\end{minted} 
		\end{minipage}
		\begin{minipage}{0.45\textwidth}
			\begin{minted}{python}
			def bfs(graph, start):
			visited = [start]
			queue = [start]
			while queue:
			node = queue.pop(0)
			for neighbor in graph[node]:
			if (neighbor not in visited):
			visited.add(neighbor)
			queue.append(neighbor)
			return visited
			\end{minted} 
		\end{minipage}
		\caption{
			\label{fig:ex_deobfuscation}
			\small
			\textbf{Full deobfuscation of a breadth-first-search function by \deobf.}
			The code on top has been fully obfuscated. The code on the bottom was recovered using \deobf by replacing the function name and every variable name using the generated dictionary. \deobf is able to suggest relevant function and variable names. It makes the code much more readable and easier to understand.
		}
	\end{figure}
	
	
	\section{Results}
	\label{sec:results}
	
	\subsection{Deobfuscation}
	\begin{figure*}[t]
		\begin{center}
			\begin{tabular}{lll}
				\toprule
				\multicolumn{1}{c}{Input Code} & \multicolumn{2}{c}{Function Name Proposals}\\
				\midrule
				\begin{minipage}{0.45\textwidth}
					\begin{minted}[escapeinside=||]{python}
					def |\colorbox{pastel_blue}{FUNC\_0}|(m1, m2):
					assert m1.shape == m2.shape
					n, m = m1.shape
					res = [[0 for _ in range(m)] for _ in range(n)]
					for i in range(n):
					for j in range(m):
					res[i][j] = m1[i][j] + m2[i][j]
					return res
					\end{minted} 
				\end{minipage}&
				{\scriptsize
					\begin{minipage}{0.2\textwidth}
						matrix\_add\\ 
						matrixAdd\\ 
						matrixadd\\ 
						matrix\_sum\\ 
						matrix\_addition \\
					\end{minipage}
				} & 
				{\scriptsize
					\begin{minipage}{0.05\textwidth}
						\score{25.944894191180843}\% \\ 
						\score{22.455118129730955}\% \\ 
						\score{18.7734687700784}\% \\ 
						\score{16.683123481678635}\% \\ 
						\score{16.14339542733116}\% \\
					\end{minipage}
				}\\ \\
				\begin{minipage}{0.45\textwidth}
					\begin{minted}[escapeinside=||]{python}
					def |\colorbox{pastel_blue}{FUNC\_0}|(matrix):
					n, _ = matrix.shape 
					for i in range(n):
					for j in range(i,n):
					matrix[i][j], matrix[j][i] = \ 
					matrix[j][i], matrix[i][j]
					\end{minted}
				\end{minipage} &
				{\scriptsize
					\begin{minipage}{0.2\textwidth}
						transpose \\ 
						rotate \\ 
						rotate\_matrix \\ 
						symmetric \\ 
						rotate\_matrix\_by\_row  \\
					\end{minipage}
				} &
				{\scriptsize
					\begin{minipage}{0.05\textwidth}
						\score{36.70172154942973}\% \\ 
						\score{29.523803383194974}\% \\ 
						\score{17.1375827019261}\% \\ 
						\score{8.898464272935005}\% \\ 
						\score{7.738428092514201}\% \\
					\end{minipage}
				} \\ \\
				
				\begin{minipage}{0.45\textwidth}
					\begin{minted}[escapeinside=||]{python}
					def |\colorbox{pastel_blue}{FUNC\_0}|(m1, m2):
					n1, m1 = m1.shape
					n2, m2 = m2.shape
					assert n2 == m1
					res = [[0 for _ in range(m2)] for _ in range(n1)]
					for i in range(n1):
					for j in range(m2):
					res[i][j] = sum([m1[i][k] * m2[k][j]
					for k in range(n2)])
					return res
					\end{minted}
				\end{minipage} &
				{\scriptsize
					\begin{minipage}{0.2\textwidth}
						matrix\_product\\ 
						mat\_mult\\ 
						matmul\_mat\\ 
						matprod\\ 
						matrixProduct \\
					\end{minipage}
				} &
				{\scriptsize
					\begin{minipage}{0.05\textwidth}
						\score{28.823747650733612}\% \\ 
						\score{23.805008872891236}\% \\ 
						\score{16.973866379559574}\% \\ 
						\score{15.952777256119369}\% \\ 
						\score{14.44459984069621}\% \\
					\end{minipage}
				} \\
				\\
				\bottomrule
			\end{tabular}
		\end{center}
		\caption{
			\label{fig:matrix_operations}
			\small
			\textbf{Additional examples of function name proposals for matrix operations in Python.}
			\deobf is able to find the right name for each matrix operation, showing that it learned to attend to the most important parts of the code. Even when the functions are similar, \deobf successfully and confidently (c.f. scores) understands the semantics of the function and its purpose.
		}
	\end{figure*}
	
	In Table~\ref{tab:res_obf}, we evaluate the ability of our model to recover identifier names, either when only one identifier is obfuscated ($p_{obf}=0$) or when all identifiers are obfuscated ($p_{obf}=1$), for models trained with $p_{obf} \in \{0, 0.5, 1\}$.
	Even when evaluating with $p_{obf}=0$, training with $p_{obf}=0$ is less efficient than $p_{obf}=0.5$ since the model is only trained to generate a single variable for each input sequence.
	Training with $p_{obf}=0.5$ is a more difficult task that requires the model to learn and understand more about code semantics.
	Forcing the model to understand the structure of the code may be useful even when testing with $p_{obf}=0$, as some identifier names cannot be guessed only from the names of other identifiers.
	When \deobf has to recover a fully obfuscated function, it obtains the best accuracy when trained with $p_{obf}=1$. It manages to recover 45.6\% of the initial identifier names.
	We also observe that, for every configuration, initializing \deobf with MLM improves the~performance.
	
	Figure~\ref{fig:ex_deobfuscation} shows an example of a fully obfuscated function recovered by our model.
	\deobf successfully manages to understand the purpose of the function and to predict appropriate variable names.
	Figure~\ref{fig:matrix_operations} shows examples of function name proposal by \deobf for functions implementing matrix operations in Python. We observe that \deobf manages to identify the key tokens and to properly infer the purpose of similar but very different functions.
	Figures~\ref{fig:examples_name_proposal_java},~\ref{fig:examples_name_proposal_python},~and~\ref{fig:vector_products} in the appendix show additional examples of function name proposals by \deobf in Java and Python.
	Figure~\ref{fig:graph_traversal} in the appendix shows additional examples where we show that \deobf also leverages non-obfuscated identifier names to understand the meaning of input functions.
	Figures~\ref{fig:deobf_lstm}~and~\ref{fig:examples_deobf_python} in the appendix show examples of deobfuscation of fully obfuscated Python code snippets using \deobf. It is able to understand the semantics and purposes of a variety of obfuscated classes and functions, including a LSTM cell.

	\begin{table}
		\caption{\small
			\label{tab:res_obf}
			\small
			\textbf{Results on partial and full deobfuscation.}
			Token accuracy and subtoken F1 score of \deobf evaluated with $p_{obf}=0$ (i.e. name proposal, where a single token is obfuscated) and $p_{obf}=1$ (i.e. full deobfuscation, where all tokens are obfuscated).
			We consider models trained with different obfuscation probabilities $p_{obf}$.
			$\text{\deobf}_{0.5}$ performs well for both tasks, and it even performs better than $\text{\deobf}_0$ for Identifier Name Proposal.
			$\text{\deobf}_0$ and $\text{\deobf}_1$ perform poorly when evaluated on other $p_{obf}$ parameters.
			Pre-training \deobf with MLM further improves the performance.
			\vspace{-0.4cm}}
		\centering
		\begin{tabular}{l ccccc}
			\\
			\toprule
			& \multicolumn{2}{c}{Eval $p_{\text{obf}}=0$} &~& \multicolumn{2}{c}{Eval $p_{\text{obf}}=1$}\\
			\cmidrule{2-3}\cmidrule{5-6}
			& Acc & F1       && Acc & F1  \\
			\midrule
			$\text{\deobf}_{0}$                   & \score{56.3272} & \score{68.0282}  && \score{0.4216} & \score{0.9138} \\
			$\text{\deobf}_{0.5}$                 & \score{61.06085} & \score{71.2274}  && \score{41.80665} & \score{54.7996} \\
			$\text{\deobf}_{1}$                 & \score{18.12775} & \score{27.004}  && \score{45.56205} & \score{58.1119} \\
			\midrule
			$\text{\deobf}_{0.5}$ init MLM                 & \textbf{\score{67.58465}} & \textbf{\score{76.3457}}  && \score{45.6549} & \score{58.0013} \\
			$\text{\deobf}_{1}$ init MLM                 & \score{20.04105} & \score{28.3414} && \textbf{\score{49.7472}} & \textbf{\score{61.1263}} \\
			
			\bottomrule
		\end{tabular}
		\vspace{-0.5cm}
	\end{table}

	\subsection{Downstream tasks}

	\vspace{-0.2cm}
	\begin{table*}[!t]
		\caption{
			\label{tab:codexglue}
			\small
			\textbf{Results on downstream tasks for different pre-training configurations.}
			Models pre-trained with \deobf initialized with MLM significantly outperform both \codebert and models trained with MLM only.
			DOBF+DAE outperforms other models on every task but clone detection, on which CodeBERT scores much higher than our MLM. 
			\resubmition{It outperforms \graphcodebert by 0.02 MRR (+5.3\%) on natural language code search (NLCS), and by 4.6\% in Java $\rightarrow$ Python computational accuracy with beam size 10 (+12.2\% correct translations).}
			The tasks where MLM provides large improvements over the transformer baseline (first row, no pre-training) are also the tasks where \deobf provides the largest gains (clone detection, NL code search, unsupervised translation).
			\newbr{The DAE baseline (initialized with MLM) already provides substantial improvements over MLM on most tasks and yields the best results for Python to Java translation while its results are poor for Java to Python.}
			\vspace{-0.4cm}
		} 
		
		\begin{center}
			\resizebox{\columnwidth}{!}{\small
				\begin{tabular}{l cccccccc|cccc}
					\\
					\toprule
					&  \multicolumn{2}{c}{Clone Det} &  \multicolumn{2}{c}{Code Sum Java} &  \multicolumn{2}{c}{Code Sum Python} &  \multicolumn{2}{c|}{NLCS} &  \multicolumn{2}{c}{Python$\rightarrow$Java} & \multicolumn{2}{c}{Java$\rightarrow$Python}\\
					
					&\multicolumn{2}{c}{(F1 score)} & \multicolumn{2}{c}{(BLEU)} & \multicolumn{2}{c}{(BLEU)} & \multicolumn{2}{c|}{(MRR)} & \multicolumn{2}{c}{(CA@1)} & \multicolumn{2}{c}{(CA@1)} \\
					
					& & & & & & & & & k=1 & k=10 & k=1 & k=10\\
					
					\midrule
					Transformer                 & \multicolumn{2}{c}{\bleu{88.14}} & \multicolumn{2}{c}{\bleu{16.5811}} & \multicolumn{2}{c}{\bleu{16.4258}} & \multicolumn{2}{c|}{\acc{0.025}} & \compacc{23.95} & \compacc{28.36} & \compacc{29.04} & \compacc{29.69}\\
					
					MLM                 & \multicolumn{2}{c}{\bleu{91.89}} & \multicolumn{2}{c}{\bleu{18.5853}} & \multicolumn{2}{c}{\bleu{17.9516}} & \multicolumn{2}{c|}{\acc{0.3076}} & \compacc{44.75} & \compacc{45.38}  & \compacc{34.5} & \compacc{35.59} \\
					DAE                 & \multicolumn{2}{c}{\bleu{96.3}} & \multicolumn{2}{c}{\bleu{19.1857}} & \multicolumn{2}{c}{\bleu{18.2796}} & \multicolumn{2}{c|}{\acc{0.3797}} 
					& \textbf{\compacc{48.32}} & \textbf{\compacc{49.16}}  & \compacc{32.1} & \compacc{32.75} \\

					\codebert                 & \multicolumn{2}{c}{\bleu{96.50}} & \multicolumn{2}{c}{\bleu{18.2526}} & \multicolumn{2}{c}{\bleu{18.2219}} & \multicolumn{2}{c|}{\acc{0.3154}} & \compacc{40.77} & \compacc{45.59} & \compacc{36.46} & \compacc{36.68}\\
					\graphcodebert{}                 & \multicolumn{2}{c}{\bleu{96.38}} & \multicolumn{2}{c}{\bleu{18.7762}} & \multicolumn{2}{c}{\bleu{18.5121}} & \multicolumn{2}{c|}{\acc{0.3767}} & \compacc{44.33} & \compacc{44.12} & \compacc{35.59} & \compacc{37.77}\\
					\hline
					\deobf init scratch        & \multicolumn{2}{c}{\textbf{\bleu{96.52}}} & \multicolumn{2}{c}{\bleu{18.1925}} & \multicolumn{2}{c}{\bleu{17.5107}} & \multicolumn{2}{c|}{\acc{0.2721}}  & \compacc{43.91} & \compacc{44.12} & \compacc{35.15} & \compacc{34.72}  \\
					
					\deobf          & \multicolumn{2}{c}{\bleu{95.87}} & \multicolumn{2}{c}{\bleu{19.0546}} & \multicolumn{2}{c}{\bleu{18.2399}}& \multicolumn{2}{c|}{\acc{0.383}} & \compacc{43.49} & \compacc{44.12}  &
					\compacc{38.65} & \compacc{39.96}\\
					\deobf{}+DAE & \multicolumn{2}{c}{\bleu{95.82}} & \multicolumn{2}{c}{\textbf{\bleu{19.3646}}} & \multicolumn{2}{c}{\textbf{\bleu{18.5767}}}& \multicolumn{2}{c|}{\textbf{\acc{0.3966}}} & \compacc{46.64} & \compacc{47.27}  & \textbf{\compacc{40.61}} & \textbf{\compacc{42.36}}\\
					\bottomrule
				\end{tabular}
			}
		\end{center}
		\vspace{-0.6cm}
	\end{table*}
	
	For fine-tuning, we considered models pre-trained with $p_{obf}=0.5$ and $p_{obf}=1$. Since they gave very similar results on downstream tasks, we only use models pre-trained with $p_{obf}=0.5$ in the rest of the paper.
	\resubmition{We initialize \deobf with MLM as it leads to better performance on our deobfuscation metrics. 
		We still consider \deobf initialized randomly as a baseline in Table~\ref{tab:codexglue}.
		We also consider a version where \deobf is trained together with a denoising auto-encoding (DAE) objective~\cite{vincent2008extracting}, which was shown to be effective at learning code representations in~\citet{roziere2020unsupervised}. 
		With DAE, the model is trained to recover the original version of a sequence which has been corrupted (by removing and shuffling tokens).
	}
	As baselines, we consider a randomly initialized model and a model pre-trained with MLM only, \newbr{and a model pre-trained with denoising and initialized with MLM.}
	For CodeXGLUE tasks, we also consider \codebert as a baseline. We compare results for \deobf trained from scratch and \deobf initialized with MLM, and report results in Table~\ref{tab:codexglue}.
	The randomly initialized model is useful to measure the importance of pre-training on a given task.
	Pre-training is particularly important for the NLCS task: without pre-training, the model achieves a performance of 0.025 MMR while it goes up to 0.308 with MLM pre-training.
	The main differences between our MLM baseline and \codebert, are that 1) \codebert was trained on a different dataset which contains functions with their documentation, 2) it uses an additional RTD objective, and 3) is initialized from a \roberta model.
	Although code summarization and NL code search involve natural language and may benefit from \codebert's dataset that contains code documentation, we obtained very similar results on this task using a simpler dataset.
	However, our MLM baseline did not match their performance on clone detection.
	We also tried to initialize our MLM model with \roberta, but did not observe any substantial impact on the performance on downstream tasks.
	
	
	\resubmition{The models based on \deobf obtain state-of-the-art results on all downstream tasks, outperforming \graphcodebert, \codebert and MLM.}
	The deobfuscation objective is already effective as a pre-training task. \resubmition{Even when initialized randomly}, it leads to results comparable to MLM on most tasks and is much more effective on clone detection.
	\resubmition{The \deobf{}+DAE model outperforms MLM on all downstream tasks, the major improvement being for NL code search, which is also the task that benefited the most from MLM pretraining}
	\resubmition{For unsupervised translation, \deobf{}+DAE increases the computational accuracy by 1.9\% when translating from Python to Java, and by 6.8\% when translating from Java to Python with beam size 10.}
	Also, \deobf beats \codebert by a wide margin on NL code search and code summarization, showing that programming language data aligned with natural language is not necessary to train an effective model on those tasks.
	\deobf initialized with MLM and combined with DAE yields higher scores than both \deobf alone and MLM, on most tasks. It shows that objectives such as MLM and DAE that provide unstructured noise are complementary to DOBF.

	\vspace{-0.1cm}
	\section{Conclusion}
	\vspace{-0.1cm}
	In this paper, we introduce a new deobfuscation objective and show that it can be used for three purposes: recover fully obfuscated code, suggest relevant identifier names, and pre-train transformer models for programming language related tasks.
	Although it does not require any parallel corpora of source code aligned to natural language, \resubmition{methods based on \deobf outperform \graphcodebert{},} \codebert and MLM pre-training on multiple downstream tasks, including clone detection, code summarization, natural language code search, and unsupervised code translation. These results show that \deobf leverages the particular structure of source code to add noise to the input sequence in a particularly effective way. 
	Other noise functions or surrogate objectives adapted to source code may improve the performance further.
	For instance, by training model to find the type of given variables, the signature of a method, or to repair a piece of code which has been corrupted.
	Since models pretrained on source code benefit from structured noise, it would be interesting to see whether these findings can be applied to natural languages as well. Although ambiguous, natural languages also have an underlying structure. Leveraging the constituency or dependency parse trees of sentences (as opposed to abstract syntax trees in programming languages) may help designing better pre-training objectives for natural languages.

	
	
	
	

	
	\bibliographystyle{plainnat}
	\bibliography{biblio}

	\appendix

	\begin{table}[h]
		\caption{\small
			\small
			\textbf{Dataset statistics.} 
			\label{tab:dataset}
		}
		
		\centering
		\begin{tabular}{l cc}
			\\
			\toprule
			& Java & Python \\
			\midrule
			All - Size & 26 GB & 19 GB \\
			All - Nb files & 7.9M & 3.6M \\
			Av. nb of tokens / file & 718 & 1245 \\ 
			Av. nb of identifiers / file & 25.9 & 41.8 \\
			\bottomrule
		\end{tabular}
		
	\end{table}
	
	\begin{figure*}[t]
		\center
		\begin{tabular}{lll}
			\toprule
			\multicolumn{1}{c}{Input Code} & \multicolumn{2}{c}{Proposed Function Name} \\
			\midrule
			\\ 
			\begin{minipage}{0.5\textwidth}
				\begin{minted}[escapeinside=||]{java}
				public static void |\colorbox{pastel_blue}{FUNC\_0}|(String path){
				try {
				Files.delete(path);
				} 
				catch (Exception e) {
				System.err.println("Error deleting file " + path);
				}
				}
				\end{minted}
			\end{minipage} & 
			\begin{minipage}{0.2\textwidth}
				{\scriptsize
					\begin{tabular}{l}
						deleteFile \\
						remove \\
						DeleteFile \\
						removeFile \\
						deleteFileQuietly  \\
				\end{tabular}}
			\end{minipage} & 
			{\scriptsize
				\begin{minipage}{0.05\textwidth}
					\score{48.341923815238154}\% \\
					\score{16.935492676755942}\% \\
					\score{13.234892458405104}\% \\
					\score{13.056936715117917}\% \\
					\score{8.43075433448288}\% \\
				\end{minipage}
			}\\
			
			\begin{minipage}{0.5\textwidth}
				\begin{minted}[escapeinside=||]{java}
				public static void |\colorbox{pastel_blue}{FUNC\_0}|(String path){
				if (!Files.exists(path)) {
				Files.createDirectories(path);
				}
				}
				\end{minted}
			\end{minipage} & 
			\begin{minipage}{0.1\textwidth}
				{\scriptsize
					\begin{tabular}{l}
						createDir \\
						createDirectory \\
						createDirIfNotExists \\
						ensureDirectoryExists \\
						createDirectoryIfNotExists \\
					\end{tabular}
				}
			\end{minipage}&
			{\scriptsize
				\begin{minipage}{0.05\textwidth}
					\score{23.5035148386157}\% \\
					\score{20.91758086886886}\% \\
					\score{20.75670685797273}\% \\
					\score{18.529933148157536}\% \\
					\score{16.29226428638517}\% \\
			\end{minipage} }\\
			
			\begin{minipage}{0.65\textwidth}
				\begin{minted}[escapeinside=||]{java}
				public static List<Pair<String, Double>> |\colorbox{pastel_blue}{FUNC\_0}|(List<String> list1, 
				List<Double> list2)
				{
				return IntStream.range(0, Math.min(list1.size(), list2.size()))
				.mapToObj(i -> new Pair<>(list1.get(i), list2.get(i)))
				.collect(Collectors.toList());
				}
				\end{minted}
			\end{minipage} & 
			\begin{minipage}{0.1\textwidth}
				
				{\scriptsize
					\begin{tabular}{l}
						zip \\ 
						intersect \\ 
						combine \\ 
						merge \\ 
						intersection \\
					\end{tabular}
				}
			\end{minipage} &
			{\scriptsize
				\begin{minipage}{0.05\textwidth}
					
					\score{28.5671420953024}\%\\
					\score{20.01981632439977}\%\\
					\score{17.942045564220376}\%\\
					\score{17.504253018299092}\%\\
					\score{15.966742997778368}\%\\
			\end{minipage} }\\
			\begin{minipage}{0.5\textwidth}
				\begin{minted}[escapeinside=||]{java}
				public static int |\colorbox{pastel_blue}{FUNC\_0}|(int n){
				int a = 0, b = 1;
				int tmp;
				for (int i = 0; i < n; i ++){
				tmp = a + b;
				a = b;
				b = tmp;
				}
				return a;
				}
				\end{minted}
			\end{minipage} & 
			\begin{minipage}{0.2\textwidth}
				
				{\scriptsize
					\begin{tabular}{l}
						fib \\ 
						fibonacci \\ 
						fibon \\ 
						fibo \\ 
						fibonacci\_series \\
					\end{tabular}
				}
			\end{minipage} &
			{\scriptsize
				\begin{minipage}{0.05\textwidth}
					\score{41.463412550689235}\% \\ 
					\score{36.59563817195719}\% \\ 
					\score{9.099137791429374}\% \\ 
					\score{8.798398132651418}\% \\ 
					\score{4.043413353272798}\% \\
			\end{minipage} }\\
			
			\begin{minipage}{0.5\textwidth}
				\begin{minted}[escapeinside=||]{java}
				public static float |\colorbox{pastel_blue}{FUNC\_0}|(List<Float> vec1,
				List<Float> vec2) {
				float size = vec1.size();
				assert size == vec2.size();
				float result = 0.0f;
				for (int i = 0; i < size; i++) {
				result += vec1.get(i) * vec2.get(i);
				}
				return result;
				}
				\end{minted}
			\end{minipage} & 
			\begin{minipage}{0.2\textwidth}
				{\scriptsize
					\begin{tabular}{l}
						dotProduct \\ 
						dot \\ 
						dot\_product \\ 
						dotproduct \\ 
						inner \\
					\end{tabular}
				}
			\end{minipage} & 
			{\scriptsize
				\begin{minipage}{0.05\textwidth}
					\score{40.85276745430576}\% \\ 
					\score{23.896996133360147}\% \\ 
					\score{16.500453396626007}\% \\ 
					\score{10.478748345901701}\% \\ 
					\score{8.271034669806385}\% \\
			\end{minipage} }\\
			\bottomrule
		\end{tabular}
		\caption{
			\label{fig:examples_name_proposal_java}
			\small
			\textbf{Examples of name proposal in Java.}
			\deobf is able to suggest relevant function names for a variety of Java methods and demonstrates its ability to understand the semantics of the code. In the first two examples, the first element in the beam shows that it is able to select relevant names in the context to find a function name: it uses \texttt{Files.delete} and \texttt{Files.createDirectories} to suggest the tokens \texttt{deleteFile} and \texttt{createDir}.
			\deobf finds relevant names for Java methods without copying any part of the other tokens. For example for the third method combining two lists as in the python \texttt{zip} function, for the fourth method which computes the n-th element of the Fibonacci series and for the last method which computes the dot product between two vectors.
		}
	\end{figure*}

	\begin{figure*}[t]
		\center
		\begin{tabular}{lll}
			\toprule
			\multicolumn{1}{c}{Input Code} & \multicolumn{2}{c}{Proposals for Highlighted Identifiers} \\
			\midrule
			\begin{minipage}{0.5\textwidth}
				\begin{minted}[escapeinside=||]{py}
				def |\colorbox{pastel_blue}{FUNC\_0}|(name):
				return os.environ[name]
				\end{minted} 
			\end{minipage} &
			\begin{minipage}{0.2\textwidth}
				{\scriptsize
					\begin{tabular}{l}
						get\_env\\ 
						get\_envvar\\ 
						env\\ 
						getenv\\ 
						get\_env\_variable \\
					\end{tabular}
				}
			\end{minipage}&
			\begin{minipage}{0.05\textwidth}
				{\scriptsize
					\begin{tabular}{l}
						\score{25.30333544749734}\% \\ 
						\score{19.322689653019555}\% \\ 
						\score{19.23088041659963}\% \\ 
						\score{18.47104388638576}\% \\ 
						\score{17.67205059649772}\% \\
					\end{tabular}
				}
			\end{minipage}\\ \\
			\begin{minipage}{0.5\textwidth}
				\begin{minted}[escapeinside=||]{python}
				def |\colorbox{pastel_blue}{FUNC\_0}|(l):
				return list(set(l))
				\end{minted}
			\end{minipage} & 
			\begin{minipage}{0.2\textwidth}
				{\scriptsize
					\begin{tabular}{l}
						unique\\ 
						remove\_duplicates\\ 
						removeDuplicates\\ 
						uniquify\\ 
						unique\_items\\
					\end{tabular}
				}
			\end{minipage} &
			\begin{minipage}{0.05\textwidth}
				{\scriptsize
					\begin{tabular}{l}
						\score{24.78403643234204}\% \\
						\score{23.83417513792686}\% \\
						\score{18.821393541801395}\% \\
						\score{18.717901646930255}\% \\
						\score{13.842493240999449}\% \\
					\end{tabular}
				}
			\end{minipage} \\ \\
			\begin{minipage}{0.5\textwidth}
				\begin{minted}[escapeinside=||]{python}
				def |\colorbox{pastel_blue}{FUNC\_0}|(path):
				with gzip.open(path, 'rb') as f:
				content = f.read()
				return content
				\end{minted}
			\end{minipage} &
			\begin{minipage}{0.2\textwidth}
				{\scriptsize
					\begin{tabular}{l}
						read\_gzip\_file\\ 
						read\_gzip\\ 
						ungzip\\ 
						gzip\_content\\ 
						gzip\_read \\
					\end{tabular}
				}
			\end{minipage} &
			\begin{minipage}{0.05\textwidth}
				{\scriptsize
					\begin{tabular}{l}
						\score{22.94622452165655}\% \\ 
						\score{22.109872398907957}\% \\ 
						\score{20.783727275807237}\% \\ 
						\score{18.17136541201951}\% \\ 
						\score{15.988810391608741}\% \\
					\end{tabular}
				}
			\end{minipage} \\  \\
			
			\begin{minipage}{0.5\textwidth}
				\begin{minted}[escapeinside=||]{python}
				def |\colorbox{pastel_blue}{FUNC\_0}(n)|: 
				v = [True for i in range(n + 1)] 
				p = 2
				while (p * p <= n): 
				if (v[p] == True): 
				for i in range(p * 2, n + 1, p): 
				v[i] = False
				p += 1
				v[0]= False
				v[1]= False
				return [p for p in range(n+1) if v[p]]
				\end{minted}
			\end{minipage} & 
			\begin{minipage}{0.2\textwidth}
				{\scriptsize
					\begin{tabular}{l}
						sieve \\
						prime\_sieve \\
						sieve\_of\_eratosthenes \\
						primes \\
						eratosthenes \\
					\end{tabular}
				}
			\end{minipage} &
			\begin{minipage}{0.05\textwidth}
				{\scriptsize
					\begin{tabular}{l}
						\score{36.142018048035485}\% \\
						\score{18.53764678714244}\% \\
						\score{15.525109742529558}\% \\
						\score{15.271315062361895}\%  \\
						\score{14.52391035993061}\% \\
					\end{tabular}
				}
			\end{minipage} \\ \\
			\begin{minipage}{0.5\textwidth}
				\begin{minted}[escapeinside=||]{python}
				def f(n): 
				|\colorbox{pastel_blue}{VAR\_0}| = [True for i in range(n + 1)] 
				p = 2
				while (p * p <= n): 
				if (|\colorbox{pastel_blue}{VAR\_0}|[p] == True): 
				for i in range(p * 2, n + 1, p): 
				|\colorbox{pastel_blue}{VAR\_0}|[i] = False
				p += 1
				|\colorbox{pastel_blue}{VAR\_0}|[0]= False
				|\colorbox{pastel_blue}{VAR\_0}|[1]= False
				return [p for p in range(n+1) if |\colorbox{pastel_blue}{VAR\_0}|[p]]
				\end{minted}
			\end{minipage} & 
			\begin{minipage}{0.2\textwidth}
				{\scriptsize
					\begin{tabular}{l}
						prime\\ 
						l\\ 
						isPrime\\ 
						a\\ 
						primes\\ 
					\end{tabular}
				}
			\end{minipage} &
			\begin{minipage}{0.05\textwidth}
				{\scriptsize
					\begin{tabular}{l}
						\score{30.569774257319965}\% \\
						\score{20.467320124850605}\% \\
						\score{18.006737153479638}\% \\
						\score{16.39155767120163}\% \\
						\score{14.56461079314816}\% \\
					\end{tabular}
				}
			\end{minipage} \\
			\bottomrule
		\end{tabular}
		\caption{\small
			\label{fig:examples_name_proposal_python}
			\textbf{Examples of name proposal in Python.} Our model trained with \deobf goes well beyond copying tokens from the context. For instance, in the first example, it understands that this function is used to get environment variables. In the second example, it proposes names related to what this function actually does (removing duplicates in a list) instead of the individual operations it uses (converting to set and then to list). The last two rows show proposals for two different identifiers in a function computing the list of prime numbers below n using the sieve of Eratosthenes. The proposals for the function name are all relevant, and the third one names exactly the algorithm which is used. The variable \texttt{v} is a list of booleans. At the end of the algorithm, \texttt{v[i]} is true if and only if \texttt{i} is prime. The proposed names \texttt{prime} and \texttt{isPrime} are very relevant as they describe what the list contains. Although \texttt{l} and \texttt{a} are not very informative, they indicate that the variable is a list or an array.}
	\end{figure*}

	\begin{figure*}[h!t]
		\begin{center}
			\begin{tabular}{lll}
				\multicolumn{1}{c}{Input Code} & \multicolumn{2}{c}{Proposed Function Name}\\
				\\
				\begin{minipage}[t]{0.45\textwidth}
					\begin{minted}[escapeinside=||]{python}
					def |\colorbox{pastel_blue}{FUNC\_0}|(v1, v2):
					assert len(v1) == len(v2)
					return [a * b for a, b in zip(v1, v2)]
					\end{minted}
				\end{minipage} &
				{\scriptsize
					\begin{minipage}{0.2\textwidth}
						\texttt{multiply\_lists }\\
						\texttt{multiply\_list }\\
						\texttt{multiply }\\
						\texttt{multiply\_vectors }\\
						\texttt{mul} 
					\end{minipage}
				}&
				{\scriptsize
					\begin{minipage}{0.05\textwidth}
						\score{28.717254652408663}\% \\ 
						\score{23.542290635036057}\% \\ 
						\score{18.092008106314882}\% \\ 
						\score{14.890852777663657}\% \\ 
						\score{14.757593828576734}\% 
					\end{minipage}
				} \\ \\
				\begin{minipage}[t]{0.5\textwidth}
					\begin{minted}[escapeinside=||]{python}
					def |\colorbox{pastel_blue}{FUNC\_0}|(v1, v2):
					assert len(v1) == len(v2)
					return sum([a * b for a, b in zip(v1, v2)])
					\end{minted}
				\end{minipage} &
				{\scriptsize
					\begin{minipage}{0.2\textwidth}
						\texttt{dotproduct }\\
						\texttt{dot\_product }\\
						\texttt{dotProduct }\\
						\texttt{dot }\\
						\texttt{multiply\_by\_addition} 
					\end{minipage}
				} &
				{\scriptsize
					\begin{minipage}{0.05\textwidth}
						\score{34.7643986373954}\% \\ 
						\score{19.231481311527844}\% \\ 
						\score{18.063236040275214}\% \\ 
						\score{15.63560230832862}\% \\ 
						\score{12.305281702472918}\% 
					\end{minipage}
				}\\ \\
				\begin{minipage}[t]{0.5\textwidth}
					\begin{minted}[escapeinside=||]{python}
					def |\colorbox{pastel_blue}{FUNC\_0}|(v1, v2):
					assert len(v1) == len(v2)
					return [a ^ b for a, b in zip(v1, v2)]
					\end{minted}
				\end{minipage} &
				{\scriptsize
					\begin{minipage}{0.2\textwidth}
						\texttt{xor}\\
						\texttt{XOR}\\
						\texttt{vector\_xor}\\
						\texttt{xors}\\
						\texttt{xor\_lists}
					\end{minipage}
				} &
				{\scriptsize
					\begin{minipage}{0.05\textwidth}
						\score{62.91804533996527}\% \\
						\score{12.829889558489882}\% \\
						\score{10.788119506139651}\% \\
						\score{7.387087503251438}\% \\
						\score{6.07685809215376}\% 
					\end{minipage}
				}\\ \\
				\begin{minipage}[t]{0.5\textwidth}
					\begin{minted}[escapeinside=||]{python}
					def |\colorbox{pastel_blue}{FUNC\_0}|(v1, v2):
					assert len(v1) == len(v2)
					return [a ** b for a, b in zip(v1, v2)]
					\end{minted}
				\end{minipage} &
				{\scriptsize
					\begin{minipage}{0.2\textwidth}
						\texttt{power} \\
						\texttt{list\_power}\\ 
						\texttt{lcm}\\ 
						\texttt{power\_list}\\ 
						\texttt{powersum}
					\end{minipage}
				} &
				{\scriptsize
					\begin{minipage}{0.05\textwidth}
						\score{29.76138156175355}\% \\
						\score{20.862327633436205}\% \\
						\score{19.923414945390565}\% \\
						\score{15.14088766467196}\% \\
						\score{14.311988194747718}\% 
					\end{minipage}
				}\\ \\
				\begin{minipage}[t]{0.5\textwidth}
					\begin{minted}[escapeinside=||]{python}
					def |\colorbox{pastel_blue}{FUNC\_0}|(v1, v2):
					assert len(v1) == len(v2)
					return [a + b for a, b in zip(v1, v2)]
					\end{minted}
				\end{minipage} &
				{\scriptsize
					\begin{minipage}{0.2\textwidth}
						\texttt{add\_lists} \\
						\texttt{add} \\
						\texttt{sum\_lists} \\
						\texttt{list\_concat} \\
						\texttt{list\_add} 
					\end{minipage}
				} &
				{\scriptsize
					\begin{minipage}{0.05\textwidth}
						\score{26.99934869782589}\% \\
						\score{22.910282783244714}\% \\
						\score{17.94884156102621}\% \\
						\score{17.678537022424383}\% \\
						\score{14.462989935478795}\% 
					\end{minipage}
				}\\ \\ 
				\begin{minipage}[t]{0.5\textwidth}
					\begin{minted}[escapeinside=||]{python}
					def |\colorbox{pastel_blue}{FUNC\_0}|(v1, v2):
					assert len(v1) == len(v2)
					return [a - b for a, b in zip(v1, v2)]
					\end{minted}
				\end{minipage} &
				{\scriptsize
					\begin{minipage}{0.2\textwidth}
						\texttt{minus}  \\
						\texttt{subtract}  \\
						\texttt{difference}  \\
						\texttt{subtract\_lists}  \\
						\texttt{substract} 
					\end{minipage}
				} &
				{\scriptsize
					\begin{minipage}{0.05\textwidth}
						\score{30.43533765531454}\% \\
						\score{29.800364444111338}\% \\
						\score{14.083504541352138}\% \\
						\score{13.290324789171201}\% \\
						\score{12.390468570050775}\% 
					\end{minipage}
				}\\
				
			\end{tabular}
		\end{center}
		\caption{\small
			\label{fig:vector_products}
			\small
			\textbf{Examples of function name proposal in Python using \deobf.}
			\deobf is able to identify the key tokens in each function, to properly infer its purpose, and to suggest appropriate names along with a confidence score.
			In particular, even though the first two code snippets are very similar in terms of edit distance, they implement very different functions and \deobf is able to name them appropriately.
		}
	\end{figure*}
	
	\begin{figure*}
		\center
		\begin{tabular}{ccc}
			\toprule
			\multicolumn{1}{c}{BFS Implementation}& \multicolumn{1}{c}{DFS Implementation} & \multicolumn{1}{c}{DFS with Erroneous} \\
			& & \multicolumn{1}{c}{Variable Name}\\
			\midrule 
			\begin{minipage}{0.3\textwidth}
				\begin{minted}[escapeinside=||]{python}
				def |\colorbox{pastel_blue}{FUNC\_0}|(graph, node):
				visited = [node]
				|\colorbox{pastel_blue}{VAR\_0}| = [node]
				while |\colorbox{pastel_blue}{VAR\_0}|:
				s = |\colorbox{pastel_blue}{VAR\_0}|.pop(0) 
				for neighbour in graph[s]:
				if neighbour not in visited:
				visited.add(neighbour)
				|\colorbox{pastel_blue}{VAR\_0}|.append(neighbour)
				return visited
				\end{minted} 
			\end{minipage} &
			\begin{minipage}{0.3\textwidth}
				\begin{minted}[escapeinside=||]{python}
				def |\colorbox{pastel_blue}{FUNC\_0}|(graph, node):
				visited = [node]
				|\colorbox{pastel_blue}{VAR\_0}| = [node]
				while |\colorbox{pastel_blue}{VAR\_0}|:
				s = |\colorbox{pastel_blue}{VAR\_0}|.pop() 
				for neighbour in graph[s]:
				if neighbour not in visited:
				visited.add(neighbour)
				|\colorbox{pastel_blue}{VAR\_0}|.append(neighbour)
				return visited
				\end{minted}
			\end{minipage} &
			\begin{minipage}{0.3\textwidth}
				\begin{minted}[escapeinside=||]{python}
				def |\colorbox{pastel_blue}{FUNC\_0}| (graph, node):
				visited = [node]
				queue = [node]
				while queue:
				s = queue.pop() 
				for neighbour in graph[s]:
				if neighbour not in visited:
				visited.append(neighbour)
				queue.append(neighbour)
				return visited
				\end{minted} 
			\end{minipage}\\
			\\
			{\scriptsize \texttt{FUNC\_0 bfs | VAR\_0 queue}}
			&
			{\scriptsize \texttt{FUNC\_0 dfs | VAR\_0 stack}}
			&
			{\scriptsize \texttt{FUNC\_0 bfs}}
			\\
			
			\bottomrule
		\end{tabular}
		\caption{\small
			\label{fig:graph_traversal}
			\textbf{Deobfuscation on graph traversal functions.} These three functions perform graph traversals. The only difference between the first and the second function is that the first uses a queue to select the next element (\texttt{.pop(0)}) while the second uses a stack (\texttt{.pop()}). The first function implements a breadth-first search (bfs) in the graph and the second implements a depth-first search (dfs). \deobf is able to find the right function and variable names in each case. 
			In the last function, we replaced the anonymized \texttt{VAR\_0} variable with \texttt{queue} in the implementation of depth-first search. This erroneous information leads \deobf to believe that this function performs breadth-first search. It shows that, just like human programmers, \deobf uses the names of the other variables to understand programs and choose relevant identifier names. When working on code with misleading identifier names, it is often preferable to obfuscate several identifiers.}
	\end{figure*}

	\begin{figure*}
		\begin{adjustbox}{width=1.05\textwidth,center}
			\begin{tabular}{cc}
				\textbf{Obfuscated Code} & \textbf{Code Deobfuscated using \deobf}\\
				\\
				\begin{minipage}{0.55\textwidth}
					\begin{minted}[escapeinside=||]{python}
					class CLASS_0(nn.Module):
					
					def __init__(VAR_0, VAR_1, VAR_2, VAR_3):
					super(CLASS_0, VAR_0).__init__()
					VAR_0.VAR_1 = VAR_1
					VAR_0.VAR_2 = VAR_2
					VAR_0.VAR_4 = nn.Linear(VAR_1, (4 * VAR_2), bias=VAR_3)
					VAR_0.VAR_5 = nn.Linear(VAR_2, (4 * VAR_2), bias=VAR_3)
					VAR_0.FUNC_0()
					
					def FUNC_0(VAR_6):
					VAR_7 = (1.0 / math.sqrt(VAR_6.VAR_8))
					for VAR_9 in VAR_6.VAR_10():
					VAR_9.data.uniform_((- VAR_7), VAR_7)
					
					def FUNC_1(VAR_11, VAR_12, VAR_13):
					(VAR_14, VAR_15) = VAR_13
					VAR_14 = VAR_14.view(VAR_14.size(1), (- 1))
					VAR_15 = VAR_15.view(VAR_15.size(1), (- 1))
					VAR_12 = VAR_12.view(VAR_12.size(1), (- 1))
					VAR_16 = (VAR_11.VAR_4(VAR_12) + VAR_11.VAR_5(VAR_14))
					VAR_17 = VAR_16[:, :(3 * VAR_11.VAR_8)].sigmoid()
					VAR_18 = VAR_16[:, (3 * VAR_11.VAR_8):].tanh()
					VAR_19 = VAR_17[:, :VAR_11.VAR_8]
					VAR_20 = VAR_17[:, VAR_11.VAR_8:(2 * VAR_11.VAR_8)]
					VAR_21 = VAR_17[:, (- VAR_11.VAR_8):]
					VAR_22 = (th.mul(VAR_15, VAR_20) + th.mul(VAR_19, VAR_18))
					VAR_23 = th.mul(VAR_21, VAR_22.tanh())
					VAR_23 = VAR_23.view(1, VAR_23.size(0), (- 1))
					VAR_22 = VAR_22.view(1, VAR_22.size(0), (- 1))
					return (VAR_23, (VAR_23, VAR_22))
					\end{minted} 
				\end{minipage}& 
				\begin{minipage}{0.55\textwidth}
					\begin{minted}[escapeinside=||]{python}
					class LSTM(nn.Module):
					
					def __init__(self, input_size, hidden_size, bias):
					super(LSTM, self).__init__()
					self.input_size = input_size
					self.hidden_size = hidden_size
					self.h1 = nn.Linear(input_size, (4 * hidden_size), bias=bias)
					self.h2 = nn.Linear(hidden_size, (4 * hidden_size), bias=bias)
					self.init_weights()
					
					def init_weights(self):
					stdv = (1.0 / math.sqrt(self.hidden_size))
					for m in self.modules():
					m.data.uniform_((- stdv), stdv)
					
					def forward(self, x, prev_state):
					(prev_h, prev_c) = prev_state
					prev_h = prev_h.view(prev_h.size(1), (- 1))
					prev_c = prev_c.view(prev_c.size(1), (- 1))
					x = x.view(x.size(1), (- 1))
					h = (self.h1(x) + self.h2(prev_h))
					s = h[:, :(3 * self.hidden_size)].sigmoid()
					c = h[:, (3 * self.hidden_size):].tanh()
					r = s[:, :self.hidden_size]
					g = s[:, self.hidden_size:(2 * self.hidden_size)]
					o = s[:, (- self.hidden_size):]
					c = (th.mul(prev_c, g) + th.mul(r, c))
					h = th.mul(o, c.tanh())
					h = h.view(1, h.size(0), (- 1))
					c = c.view(1, c.size(0), (- 1))
					return (h, (h, c))
					\end{minted}
				\end{minipage} \\ 
			\end{tabular}
		\end{adjustbox} 
		\\
		\\
		\begin{center}
			{\scriptsize
				\begin{tabular}{lll}
					\toprule
					ID &  Ground Truth & \deobf\\ 
					\midrule
					\texttt{CLASS\_0} & \texttt{LSTM} & \texttt{LSTM} \\
					\texttt{FUNC\_0} & \texttt{reset\_parameters} & \texttt{init\_weights} \\
					\texttt{FUNC\_1} & \texttt{forward} & \texttt{forward} \\
					\texttt{VAR\_0} & \texttt{self} & \texttt{self} \\
					\texttt{VAR\_1} & \texttt{input\_size} & \texttt{input\_size} \\
					\texttt{VAR\_2} & \texttt{hidden\_size} & \texttt{hidden\_size} \\
					\texttt{VAR\_3} & \texttt{bias} & \texttt{bias} \\
					\texttt{VAR\_4} & \texttt{i2h} & \texttt{h1} \\
					\texttt{VAR\_5} & \texttt{h2h} & \texttt{h2} \\
					\texttt{VAR\_6} & \texttt{self} & \texttt{self} \\
					\texttt{VAR\_7} & \texttt{std} & \texttt{stdv} \\
					\texttt{VAR\_8} & \texttt{hidden\_size} & \texttt{hidden\_size} \\
					\texttt{VAR\_9} & \texttt{w} & \texttt{m} \\
					\texttt{VAR\_10} & \texttt{parameters} & \texttt{modules} \\
					\texttt{VAR\_11} & \texttt{self} & \texttt{self} \\
					\texttt{VAR\_12} & \texttt{x} & \texttt{x} \\
					\texttt{VAR\_13} & \texttt{hidden} & \texttt{prev\_state} \\
					\texttt{VAR\_14} & \texttt{h} & \texttt{prev\_h} \\
					\texttt{VAR\_15} & \texttt{c} & \texttt{prev\_c} \\
					\texttt{VAR\_16} & \texttt{preact} & \texttt{h} \\
					\texttt{VAR\_17} & \texttt{gates} & \texttt{s} \\
					\texttt{VAR\_18} & \texttt{g\_t} & \texttt{c} \\
					\texttt{VAR\_19} & \texttt{i\_t} & \texttt{r} \\
					\texttt{VAR\_20} & \texttt{f\_t} & \texttt{g} \\
					\texttt{VAR\_21} & \texttt{o\_t} & \texttt{o} \\
					\texttt{VAR\_22} & \texttt{c\_t} & \texttt{c} \\
					\texttt{VAR\_23} & \texttt{h\_t} & \texttt{h} \\
					\bottomrule
				\end{tabular}
			}
		\end{center}
		\caption{
			\label{fig:deobf_lstm}
			\small
			\textbf{Deobfuscation of an LSTM cell.}
			\deobf is able to recover several of the original tokens, including the class name (\texttt{LSTM}) and the full signature of the \texttt{\_\_init\_\_} method. Even though \deobf does not always recover the original token, it generally proposes very relevant tokens which improves code readability. In particular, for some tokens the accuracy and subtoken scores would be zero but the recovered tokens are still very relevant. For instance, \texttt{reset\_parameters} (\texttt{FUNC\_0}) was renamed to \texttt{init\_weights}, \texttt{std} (\texttt{VAR\_7}) was renamed to \texttt{stdv}, and \texttt{hidden} (\texttt{VAR\_13}) was renamed to \texttt{prev\_state}. In those instances, the original and recovered tokens share no subtoken despite having very similar semantics.
		}
	\end{figure*}

	\begin{figure*}[t]
		\center
		\begin{tabular}{lll}
			\toprule
			\multicolumn{1}{c}{Input Code} & \multicolumn{2}{c}{Deobfuscated Identifiers} \\
			\midrule
			\\
			\begin{minipage}{0.5\textwidth}
				\begin{minted}{python}
				def FUNC_0(VAR_0, VAR_1):
				return sum(map(operator.mul, VAR_0, VAR_1))
				\end{minted}
			\end{minipage} & 
			\begin{minipage}{0.1\textwidth}
				{\scriptsize
					\begin{tabular}{l}
						FUNC\_0\\ 
						VAR\_0\\ 
						VAR\_1\\ 
					\end{tabular}
			} \end{minipage}&
			\begin{minipage}{0.1\textwidth}
				{\scriptsize
					\begin{tabular}{l}
						dotProduct \\ 
						list1 \\ 
						list2 \\ 
					\end{tabular}
				}
			\end{minipage}\\ \\
			
			\begin{minipage}{0.5\textwidth}
				\begin{minted}{python}
				def FUNC_0(VAR_0):
				VAR_1 = urllib2.urlopen(VAR_0)
				VAR_2 = VAR_1.read()
				return VAR_2
				\end{minted}
			\end{minipage} & 
			\begin{minipage}{0.1\textwidth}
				{\scriptsize
					\begin{tabular}{l}
						FUNC\_0\\ 
						VAR\_0\\ 
						VAR\_1\\ 
						VAR\_2\\ 
					\end{tabular}
			} \end{minipage}&
			\begin{minipage}{0.1\textwidth}
				{\scriptsize
					\begin{tabular}{l}
						get\_html \\ 
						url \\ 
						response \\ 
						html \\
					\end{tabular}
				}
			\end{minipage}\\ \\
			
			\begin{minipage}{0.5\textwidth}
				\begin{minted}{python}
				def FUNC_0(VAR_0):
				VAR_1 = set(VAR_0)
				return (len(VAR_1) == len(VAR_0))
				\end{minted}
			\end{minipage} & 
			\begin{minipage}{0.1\textwidth}
				{\scriptsize
					\begin{tabular}{l}
						FUNC\_0\\ 
						VAR\_0\\ 
						VAR\_1\\ 
					\end{tabular}
			} \end{minipage}&
			\begin{minipage}{0.1\textwidth}
				{\scriptsize
					\begin{tabular}{l}
						all\_unique \\ 
						iterable \\ 
						s \\
					\end{tabular}
				}
			\end{minipage}\\ \\

			\begin{minipage}{0.5\textwidth}
				\begin{minted}{python}
				def FUNC_0(VAR_0, VAR_1):
				return list(collections.deque(VAR_0, maxlen=VAR_1))
				\end{minted}
			\end{minipage} & 
			\begin{minipage}{0.1\textwidth}
				{\scriptsize
					\begin{tabular}{l}
						FUNC\_0\\ 
						VAR\_0\\ 
						VAR\_1\\ 
					\end{tabular}
			} \end{minipage}&
			\begin{minipage}{0.1\textwidth}
				{\scriptsize
					\begin{tabular}{l}
						tail\\ 
						s \\ 
						n \\
					\end{tabular}
				}
			\end{minipage}\\ \\
			
			\begin{minipage}{0.6\textwidth}
				\begin{minted}{python}
				def FUNC_0(VAR_0):
				return sum((VAR_1 for VAR_1 in VAR_0 if ((VAR_1 % 2) == 0)))
				\end{minted}
			\end{minipage} & 
			\begin{minipage}{0.1\textwidth}
				{\scriptsize
					\begin{tabular}{l}
						FUNC\_0\\ 
						VAR\_0\\ 
						VAR\_1\\ 
					\end{tabular}
			} \end{minipage}&
			\begin{minipage}{0.1\textwidth}
				{\scriptsize
					\begin{tabular}{l}
						even\_sum\\ 
						nums \\ 
						n \\
					\end{tabular}
				}
			\end{minipage}\\ \\
			\bottomrule
		\end{tabular}
		\caption{
			\label{fig:examples_deobf_python}
			\small
			\textbf{Examples of full deobfuscations of Python functions.}
			Even when every identifier is obfuscated, \deobf is able to propose relevant names. The proposed function name is informative and relevant in all examples since the first function computes a dot product, the second downloads a HTML page and returns its content, the third evaluates whether the input contains only unique elements, the fourth computes the tail of an iterable, and the fifth computes the sum of the even elements of an iterable.
		}
	\end{figure*}

	\begin{table}[!t]
		\caption{
			\label{tab:codexglue_transcoder}
			\small
			\textbf{Results on downstream tasks with the architecture of \transcoder.}
			This architecture has less layers (6 instead of 12), a higher embedding dimension (1024 instead of 768) and less activation heads (8 instead of 12) resulting in a slightly larger model (143M parameters instead of 126M). It also uses reLU activations instead of geLU.
			Models pre-trained with MLM and \deobf significantly outperform both \codebert and models trained with MLM only.
			MLM+\deobf outperforms \codebert by 7\% on natural language code search (NLCS), and MLM by 6\% in Java $\rightarrow$ Python computational accuracy.
			It also beats CodeBERT on every task except Clone Detection, on which CodeBERT scores much higher than our MLM. 
			\resubmition{\graphcodebert only beats our model on python summarization and Python to Java translation by a shallow margin and is below on other tasks.}
			The tasks where MLM provides large improvements over the transformer baseline (first row) are also those where \deobf provides the largest gains (i.e.~clone detection, natural language code search, and unsupervised~translation).
		} 
		\centering
		\resizebox{\columnwidth}{!}{
			\begin{tabular}[ht]{l cccccccc|cccc}
				\\
				\toprule
				&  \multicolumn{2}{c}{Clone Det} &  \multicolumn{2}{c}{Sum Java} &  \multicolumn{2}{c}{Sum Py} &  \multicolumn{2}{c|}{NLCS} &  \multicolumn{2}{c}{Py$\rightarrow$Ja} & \multicolumn{2}{c}{Ja$\rightarrow$Py}\\
				
				&\multicolumn{2}{c}{(F1 score)} & \multicolumn{2}{c}{(BLEU)} & \multicolumn{2}{c}{(BLEU)} & \multicolumn{2}{c|}{(MRR)} & \multicolumn{2}{c}{(CA@1)} & \multicolumn{2}{c}{(CA@1)} \\
				
				& & & & & & & & & k=1 & k=10 & k=1 & k=10\\
				
				\midrule
				{\small Transformer}                 & \multicolumn{2}{c}{\bleu{88.14}} & \multicolumn{2}{c}{\bleu{16.5811}} & \multicolumn{2}{c}{\bleu{16.4258}} & \multicolumn{2}{c|}{\acc{0.025}} & \compacc{37.6} & \compacc{38.8889} & \compacc{31.75} & \compacc{42.1166}\\
				
				\codebert                 & \multicolumn{2}{c}{\bleu{96.50}} & \multicolumn{2}{c}{\bleu{18.2526}} & \multicolumn{2}{c}{\bleu{18.2219}} & \multicolumn{2}{c|}{\acc{0.3154}} & - & - & - & -\\
				{\small\graphcodebert{}}                 & \multicolumn{2}{c}{\bleu{96.38}} & \multicolumn{2}{c}{\bleu{18.7762}} & \multicolumn{2}{c}{\textbf{\bleu{18.5121}}} & \multicolumn{2}{c|}{\acc{0.3767}} & - & - & - & -\\
				{\small MLM}                 & \multicolumn{2}{c}{\bleu{91.89}} & \multicolumn{2}{c}{\bleu{18.5853}} & \multicolumn{2}{c}{\bleu{17.9516}} & \multicolumn{2}{c|}{\acc{0.3076}} & \compacc{40.3} & \compacc{42.2}  & \compacc{44.7} & \compacc{46.6} \\
				
				\hline
				{\small\deobf}                & \multicolumn{2}{c}{\textbf{\bleu{96.52}}} & \multicolumn{2}{c}{\bleu{18.1925}} & \multicolumn{2}{c}{\bleu{17.5107}} & \multicolumn{2}{c|}{\acc{0.2721}}  & \compacc{38.8773} & \textbf{\compacc{45.738}} & \compacc{44.7} & \compacc{46.4363}  \\
				
				{\small MLM+\deobf}          & \multicolumn{2}{c}{\bleu{95.87}} & \multicolumn{2}{c}{\textbf{\bleu{19.0546}}} & \multicolumn{2}{c}{\bleu{18.2399}}& \multicolumn{2}{c|}{\textbf{\acc{0.383}}} & \textbf{\compacc{43.45}} & \compacc{44.9}  & \textbf{\compacc{49.24}} & \textbf{\compacc{52.488}}\\
				\bottomrule
			\end{tabular}
		}
	\end{table}
\end{document}